\documentclass{article}

\usepackage{PRIMEarxiv}

\usepackage[utf8]{inputenc} 
\usepackage[T1]{fontenc}    
\usepackage{hyperref}       
\usepackage{url}            
\usepackage{booktabs}       
\usepackage{amsfonts}       
\usepackage{nicefrac}       
\usepackage{microtype}      
\usepackage{lipsum}
\usepackage{fancyhdr}       
\usepackage{graphicx}       
\graphicspath{{media/}}     
\usepackage{graphicx}
\usepackage{float}
\usepackage{caption}

\title{The Opportunities and Risks of Large Language Models in Mental Health
\thanks{\textit{\underline{Citation}}: 
\textbf{Lawrence HR, Schneider RA, Rubin SB, Matarić MJ, McDuff DJ, Jones Bell M. The Opportunities and Risks of Large Language Models in Mental Health
JMIR Ment Health 2024;11:e59479
doi: 10.2196/59479}} 
}

\author{
  Hannah R Lawrence \\
  Google via Magnit \\
Folsom, CA, United States\\
  \texttt{hannahlawrence@google.com} \\
   \And
  Renee A Schneider \\
  Google \\
Mountain View, CA, United States\\
   \And
Susan B Rubin \\
  Google via Magnit \\
Folsom, CA, United States\\
   \And
  Maja J Matarić \\
  Google \\
Mountain View, CA, United States\\
   \And
 Daniel J McDuff \\
  Google \\
Mountain View, CA, United States\\
   \AND
 Megan Jones Bell \\
  Google \\
Mountain View, CA, United States\\
  \texttt{meganjonesbell@google.com} \\
}

\begin{document}
\maketitle

\begin{abstract}
Global rates of mental health concerns are rising, and there is increasing realization that existing models of mental health care will not adequately expand to meet the demand. With the emergence of large language models (LLMs) has come great optimism regarding their promise to create novel, large-scale solutions to support mental health. Despite their nascence, LLMs have already been applied to mental health–related tasks. In this paper, we summarize the extant literature on efforts to use LLMs to provide mental health education, assessment, and intervention and highlight key opportunities for positive impact in each area. We then highlight risks associated with LLMs’ application to mental health and encourage the adoption of strategies to mitigate these risks. The urgent need for mental health support must be balanced with responsible development, testing, and deployment of mental health LLMs. It is especially critical to ensure that mental health LLMs are fine-tuned for mental health, enhance mental health equity, and adhere to ethical standards and that people, including those with lived experience with mental health concerns, are involved in all stages from development through deployment. Prioritizing these efforts will minimize potential harms to mental health and maximize the likelihood that LLMs will positively impact mental health globally.
\end{abstract}


\section{Introduction}
Globally, half of all individuals will experience a mental health disorder in their lifetimes [1], and at any given point, 1 in 8 people are experiencing a mental health concern [2]. Despite greater attention provided in the recent years to mental health, the rate of mental health concerns has increased [2,3], and access to mental health care has not expanded to adequately meet the demand [4]. In the United States alone, the average time between the onset of mental health symptoms and treatment is 11 years [5], and nearly half of the global population lives in regions with a shortage of mental health professionals [2].

To overcome inadequate access to effective and equitable mental health care, large-scale solutions are needed. The emergence of large language models (LLMs) brings hope regarding their application to mental health and their potential to provide such solutions due to their relevance to mental health education, assessment, and intervention. LLMs are artificial intelligence models trained using extensive data sets to predict language sequences [6]. By leveraging huge neural architectures, LLMs can organize complex and abstract concepts. This enables them to identify, translate, predict, and generate new content. LLMs can be fine-tuned for specific domains (eg, mental health) and enable interactions in natural language, as do many mental health assessments and interventions, highlighting the enormous potential they have to revolutionize mental health care. In this paper, we first summarize the research done to date applying LLMs to mental health. Then, we highlight key opportunities and risks associated with mental health LLMs and put forth suggested risk mitigation strategies. Finally, we make recommendations for the responsible use of LLMs in the mental health domain.

\section{Applications of LLMs to Mental Health}
\label{sec:headings}
\subsection{Overview}

Initial tests of LLMs’ capabilities across mental health education, assessment, and intervention are promising. When considering this literature base, which we review next, it is important to first distinguish between general-purpose, consumer LLMs (eg, ChatGPT [OpenAI] and Gemini [Google]) and domain-specific LLMs (eg, Med-LM [Google]). General-purpose LLMs are trained on large corpora of text and are designed to perform a wide range of tasks. Domain-specific LLMs, on the other hand, typically build upon general-purpose LLMs through various strategies of fine-tuning with curated data to complete tasks within an area of focus. Given that general-purpose LLMs are largely trained with unrestricted text, they risk generating inaccurate, biased, stigmatizing, and harmful information about mental health. Developers of domain-specific LLMs can mitigate some of this risk by incorporating strategies during fine-tuning and evaluation such as using high-quality evidence-based information and attribution techniques [7], but it remains difficult to remove all possible risk from LLM-generated content. Given these important distinctions, in the paper that follows we clarify when findings are specific to general-purpose versus domain-specific LLMs where possible.

\subsection{Education}
One area of opportunity for LLMs in the mental health domain is to provide education about mental health (see Figure 1) [8]. Although lagging behind the success of LLMs in the medical domain [9], there is evidence that LLMs are capable of generating accurate, helpful, and immediate mental health information. The psychological support with LLM (Psy-LLM), for example, is a domain-specific LLM designed to answer mental health questions [10]. Psy-LLM was pretrained with a data set of psychology articles, question-answer pairs from psychologists, and by crawling social media platforms. The model achieved moderate levels of helpfulness, fluency, relevance to the question asked, and logic based on human ratings of Psy-LLM responses.

The abilities of general-purpose LLMs to answer questions about mental health has also been evaluated. Sezgin et al [11] compared Google Search, GPT-4 (using ChatGPT), and LaMDA (using Bard [Google DeepMind]) responses to questions about postpartum depression relative to responses from an American College of Obstetricians and Gynecologists (ACOG) frequently asked questions document. Board-certified human physicians rated ChatGPT responses as more in line with ACOG responses than Bard or Google Search responses, and on average, ChatGPT responses were rated at near ceiling for clinical accuracy, scoring a 3.93 out of a possible 4. Importantly, however, general-purpose LLMs differ in their policies regarding the generation of medical or mental health advice. Bard’s accuracy ratings were impacted by Bard’s policy to advise consulting a health care provider when asked questions about mental health. This practice protects individuals from potential harm, though such responses received lower ratings of quality in this study.

LLM-generated answers to mental health questions may not be comparable to human-generated answers, however. It is critical for LLMs to meet or exceed human performance in order for LLMs to be trusted and to ease the demand for human providers. In the case of Psy-LLM and ChatGPT, there is evidence that responses to mental health and substance use questions fall short of human-generated responses in dimensions such as accuracy, quality, and alignment with evidence-based practice (EBP) [10,12].

Another way that LLMs may serve to educate is to support provider training. Barish et al [13] used ChatGPT to generate content and associated learning objectives for an online learning platform for behavioral health professionals. Researchers compared the time providers needed to write their own content versus the time needed to edit ChatGPT-generated content, finding that using ChatGPT improved provider efficiency by 37.5 percent. LLMs can also be leveraged to train providers to optimize interactions with their patients. As two examples, Chan and Li [14] developed a chatbot trained to mimic a patient capable of describing their mental health symptoms in colloquial terms, and Sharma et al [15] used artificial intelligence to coach peer support providers to increase empathetic responding. These approaches illustrate ways that LLMs can support provider training and potentially enhance provider efficacy without providers becoming reliant on LLMs for in the moment critical thinking or decision-making.

\begin{figure}[H]
  \centering
  \includegraphics[width=0.8\textwidth]{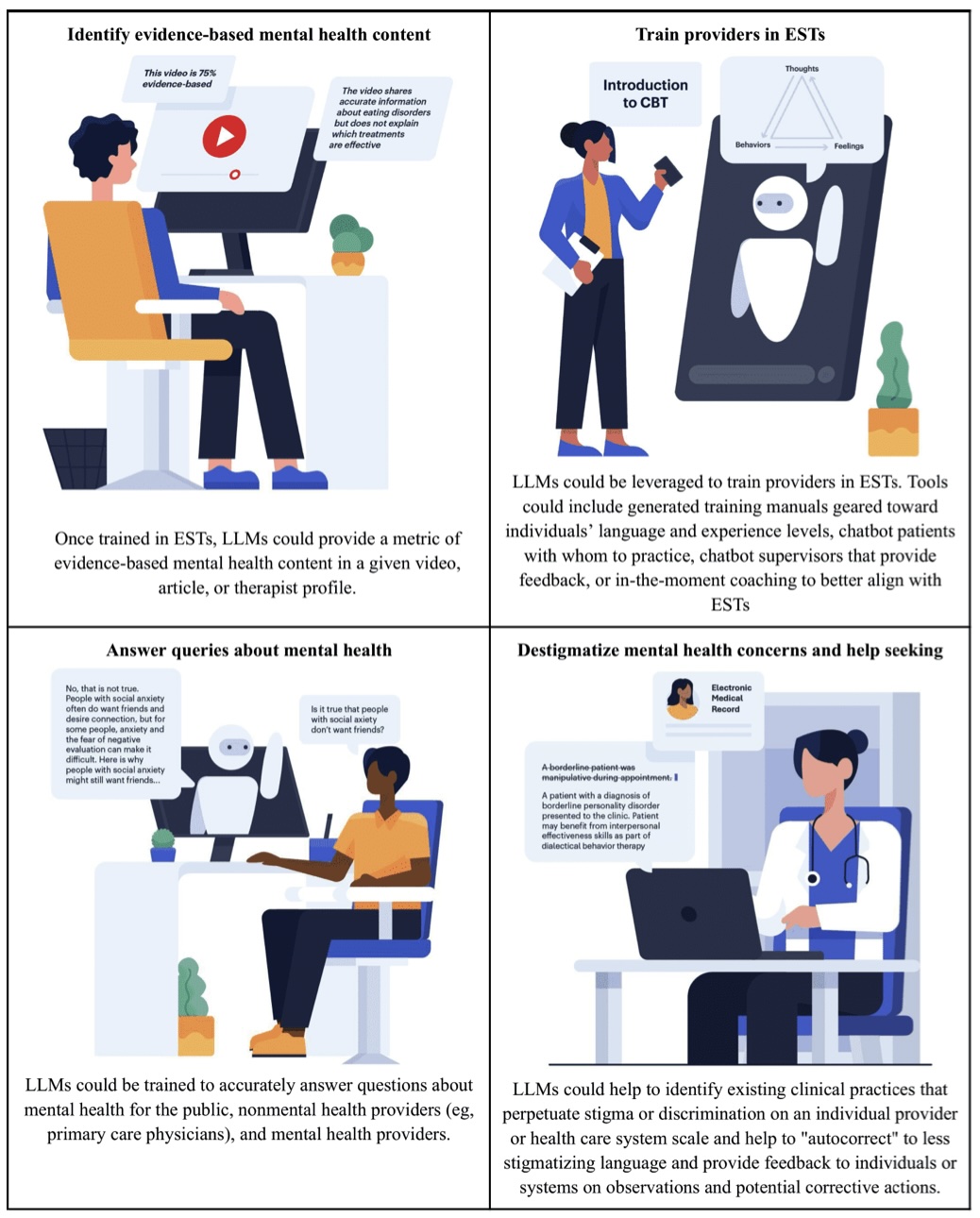}
  \caption{Potential opportunities for LLMs in mental health education. CBT: cognitive behavioral therapy; EST: empirically supported treatment; LLM: large language model.}
  \label{fig:myimage1}
\end{figure}

\subsection{Assessment}

A second function of LLMs within the domain of mental health is to assess mental health symptoms, identify diagnoses, and track changes in mental well-being (see Figure 2). LLMs can at times predict mental health symptoms and diagnoses accurately. Ji et al [16] initially developed two domain-specific models, MentalBERT and MentalRoBERTa, pretrained on mental health information. Compared with existing models pretrained in different domains, specifically clinical notes and biomedicine, MentalBERT and MentalRoBERTa were generally better able to detect depression and suicidal ideation from social media posts (notably, these results were achieved with Bidirectional Encoder Representations From Transformers [BERT]-based models that represent early-genderation LLMs, with newer models and architectures demonstrating potential for even more advanced capabilities). LLMs such as Mental-Alpaca, a mental health domain–specific LLM, Med-PaLM 2, a medical domain–specific LLM, and ChatGPT, which is general-purpose, have also been shown to screen for possible depressive symptoms and suicide risk, with varying degrees of accuracy [17-20].

When it comes to predicting mental health diagnoses specifically, there is evidence that Med-PaLM 2 can do so accurately. When presented with a series of case studies from the American Psychiatric Association book of DSM-5 (Diagnostic and Statistical Manual of Mental Disorders, Fifth Edition) case examples [21], Med-PaLM 2 predicted the correct diagnosis 77.5 percent of the time, and performance increased to 92.5 percent when asked to specify the correct diagnostic category (eg, depressive disorder vs major depressive disorder) [20]. Similarly, when PaLM 2 was fine-tuned with medical domain data and optimized for differential diagnosis, the model was able to generate more appropriate and comprehensive lists of diagnoses than specialist medical doctors in response to challenging case studies, some of which involved psychiatric diagnoses [22].

LLM-predicted assessments do not, however, always match those of human mental health clinicians, suggesting that more work is needed before LLMs can engage in assessment without human oversight. In one study [23], four iterations of a case vignette [24] were presented to ChatGPT. Each vignette varied in levels of perceived burdensomeness and thwarted belongingness—two primary risk factors for suicide [25,26]. ChatGPT appropriately determined that the risk for suicidal ideation and suicide attempts was highest for the vignette with both high perceived burdensomeness and high thwarted belongingness, but it predicted lower suicide risk overall than did mental health professionals who reviewed the same vignettes. Med-PaLM 2 also at times does not achieve human clinician-level performance. The model predicted more severe posttraumatic stress disorder symptoms than human clinicians from clinical interview data, classified possible cases of posttraumatic stress disorder with high specificity (0.98) but low sensitivity (0.30), and the model only correctly predicted whether a case example had a comorbid diagnosis or diagnostic modifier 20 percent of the time [20].

In all the efforts described thus far, LLMs had been provided with information about symptoms and tasked with determining whether those symptoms indicated a possible mental health concern or diagnosis. LLMs also may be leveraged to ask the questions needed to screen for a mental health concern or to predict a mental health diagnosis. Chan and Li [14] developed a chatbot trained to engage in mental health assessment with patients. Compared with human psychiatrists, the chatbot displayed more empathy and asked more thorough questions about some symptoms (eg, sleep), but was less likely to rule out associated conditions.

\begin{figure}[H]
  \centering
  \includegraphics[width=0.8\textwidth]{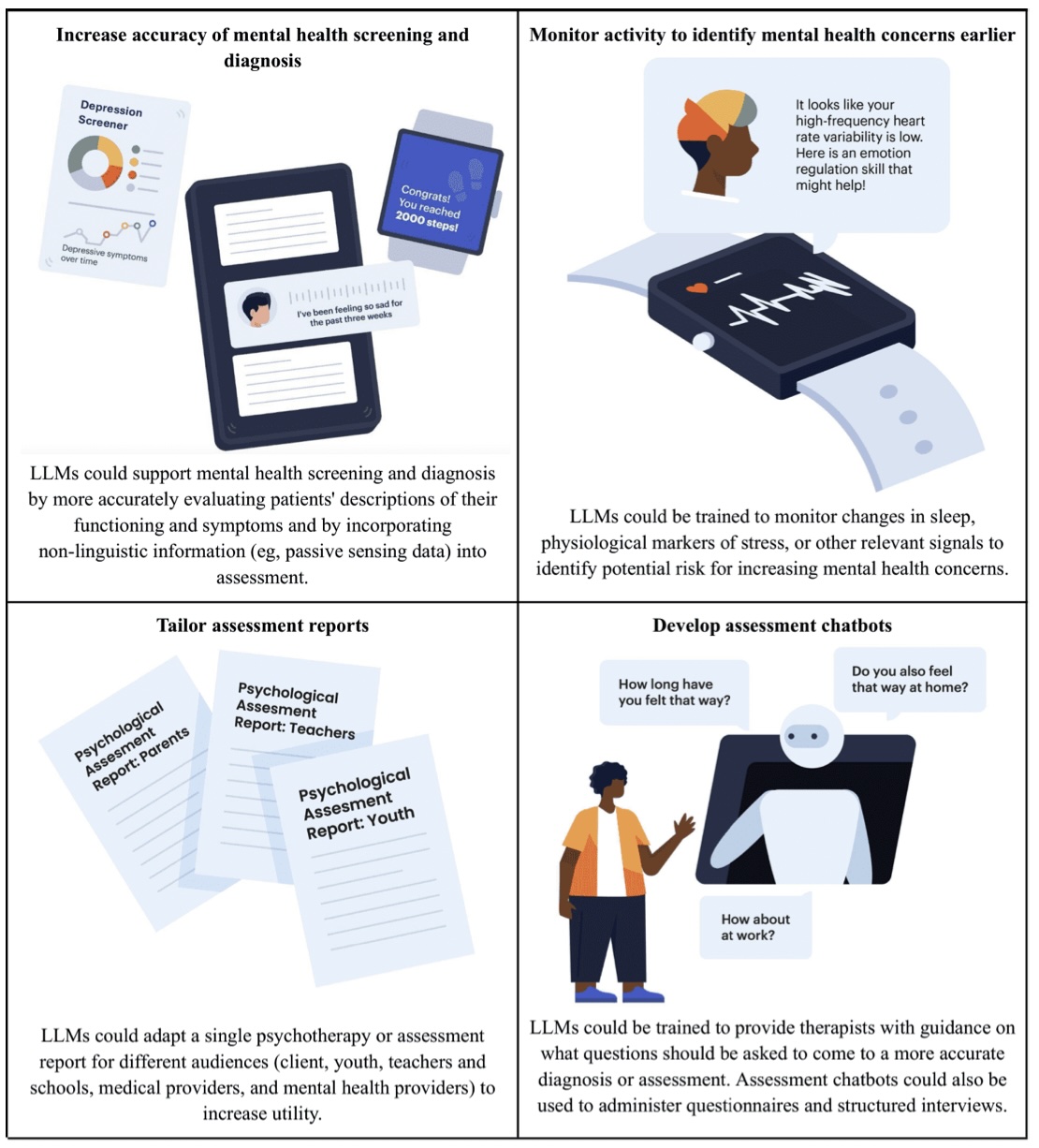}
  \caption{Potential opportunities for LLMs in mental health assessment. LLM: large language model.}
  \label{fig:myimage2}
\end{figure}

\subsection{Intervention}

A third opportunity for LLMs in the mental health domain is to implement mental health interventions (see Figure 3). To date, such efforts have largely focused on chatbots. Prominent chatbots, some of which are LLM-based, include Woebot [27], Wysa [28], Tess [29], Replika [30], Ellie [31], and Sibly [32]. Many of these chatbots were trained in empirically supported treatments such as cognitive behavioral therapy, dialectical behavior therapy, and motivational interviewing. There is initial evidence that such chatbots may be effective in reducing depressive and anxiety symptoms, as well as stress [33-36]. Additionally, research finds that chatbots can be trained to express empathy [37-39], provide nonjudgmental responses [40], and maintain therapeutic conversations [14] and that individuals can establish therapeutic rapport with chatbots [41].

Caution is warranted when using chatbots to deliver mental health interventions. To date, chatbots are not effective in treating all types of mental health distress [36] and at times have difficulty personalizing interventions [38], forget information (eg, that they had talked with someone previously) [37], and provide nontherapeutic and iatrogenic advice including encouraging substance use, dieting, and weight loss [40,42,43]. Also concerning is that chatbots do not consistently or adequately respond to suicide risk, at times being dismissive and neglecting to provide crisis resources or referals to human providers [38,44].

\begin{figure}[h!]
  \centering
  \includegraphics[width=0.8\textwidth]{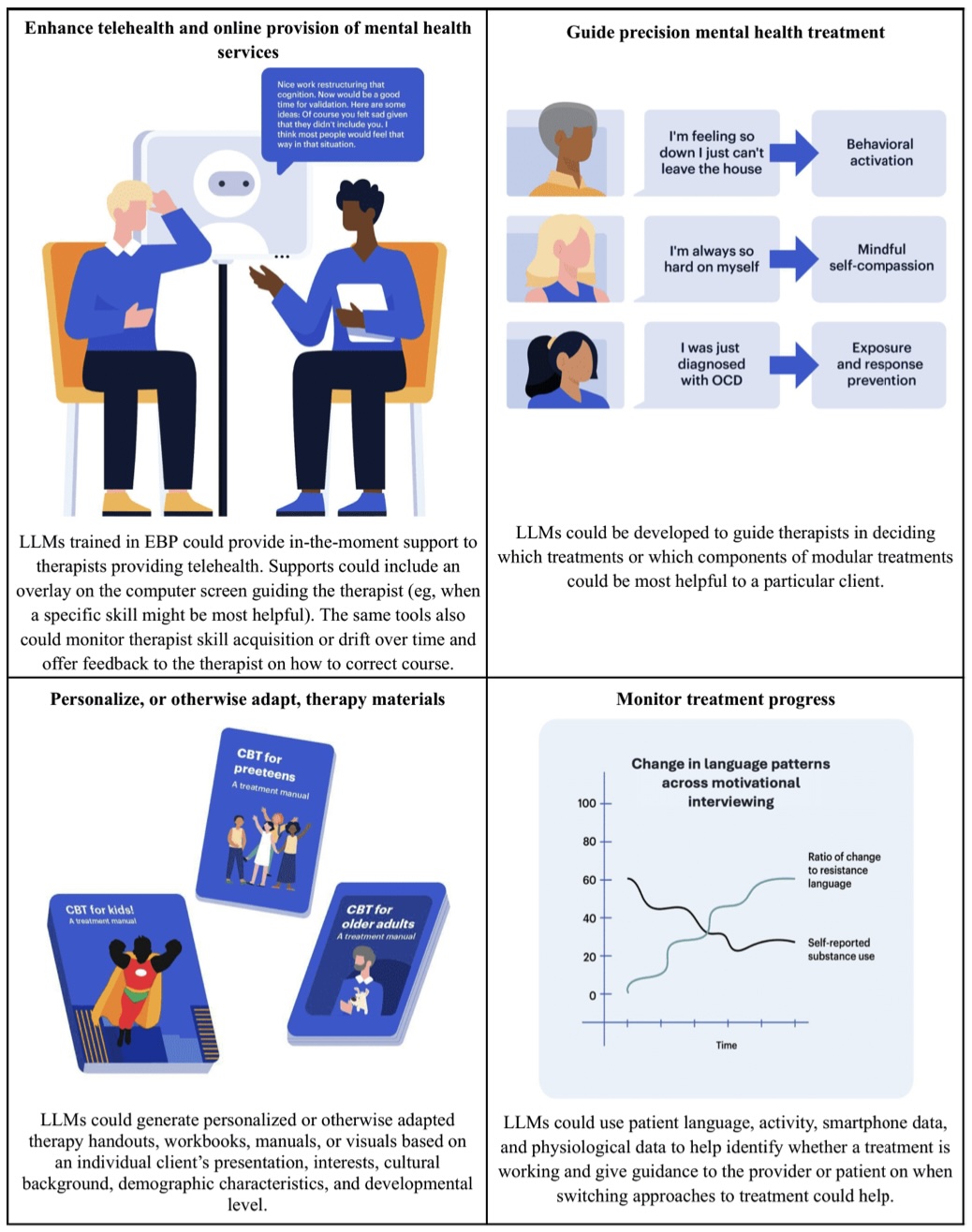}
  \caption{Potential opportunities for LLMs in mental health intervention. CBT: cognitive behavioral therapy; EBP: evidence-based practice; LLM: large language model.}
  \label{fig:myimage3}
\end{figure}

\section{Risks Associated With Mental Health LLMs}
\label{sec:headings}
\subsection{Overview}

To maximize the positive impact of LLMs on mental health, LLM development, testing, and deployment must be done ethically and responsibly (see Textbox 1). This requires identification and evaluation of risks, taking preemptive steps to mitigate risks, and establishing plans to monitor for ongoing or new and unexpected risks [45,46]. It is also important to recognize that the risks associated with the use of LLMs for mental health support may differ across education, assessment, and intervention (see Table 1). Here, we highlight primary risks that largely cut across uses of LLMs for mental health-related tasks and identify potential steps that can be taken to mitigate these risks.

\begin{figure}[htbp]
  \centering
  \includegraphics[width=0.8\textwidth]{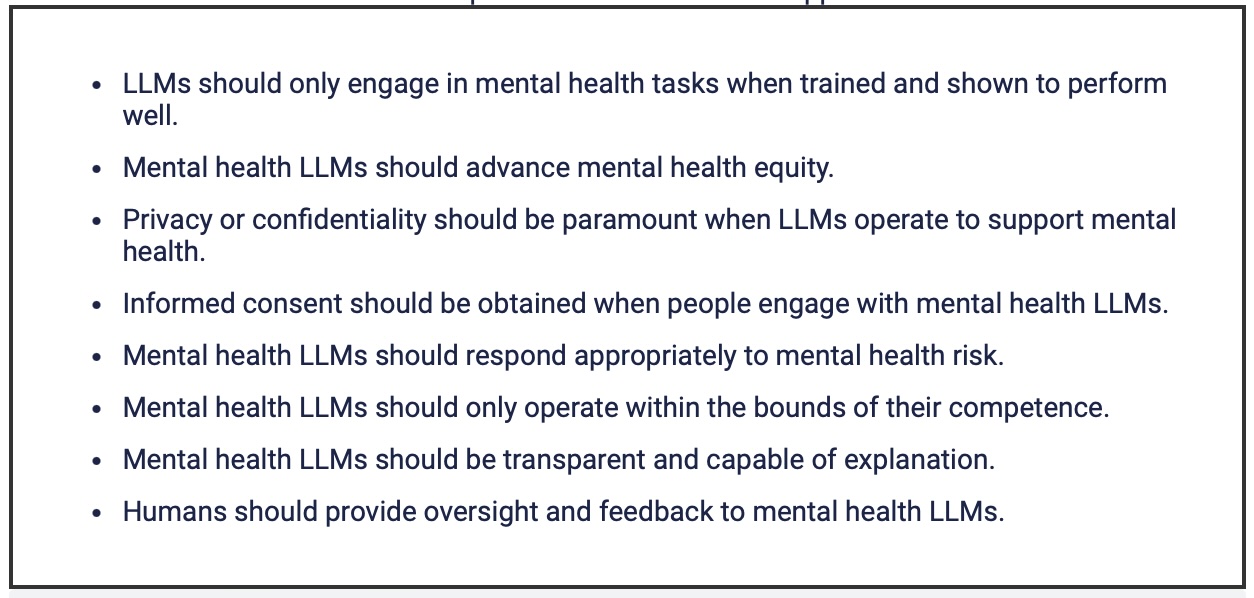}
  \vspace{0.5cm} 
  \caption{Textbox1: Recommendations for responsible use of LLMs to support mental health.}
  \label{fig:myimage3}
\end{figure}

\begin{table}[htbp]
\centering
\begin{tabular}{l|p{2.5cm}p{2.5cm}p{2.5cm}}
\toprule
Risk & Mental health education & Mental health assessment & Mental health intervention \\
\midrule
Perpetuate inequalities, disparities, & Medium & Higher & Higher \\
and stigma & & & \\
Unethical provision of mental health & & & \\
services & & & \\
\hspace{1cm} Practice beyond the & Lower & Higher & Higher \\
\hspace{1cm} boundaries of competence & & & \\
\hspace{1cm} Neglect to obtain informed & Lower & Higher & Higher \\
\hspace{1cm} consent & & & \\
\hspace{1cm} Fail to preserve \newline confidentiality & Lower & Higher & Higher \\
\hspace{1cm} or privacy & & & \\
\hspace{1cm} Build and maintain \newline inappropriate & Lower & Medium & Higher \\
\hspace{1cm} levels of trust & & & \\
Lack reliability & Lower & Higher & Higher \\
Generate inaccurate or iatrogenic & Medium & Higher & Higher \\
output & & & \\
Lack transparency or explainability & Lower & Medium & Medium \\
Neglect to involve humans & Lower & Medium & Higher \\
\bottomrule
\end{tabular}
\caption{Potential risks to people when LLMs engage in mental health education, assessment, and intervention.}
\label{my_table}
\end{table}

\subsection{Perpetuating Inequalities, Disparities, and Stigma}

There exists the risk that LLMs perpetuate inequities and stigma, further widening mental health disparities [47]. Mental health concerns are highly stigmatized [48], and there are disparities in who is at risk for mental health concerns, in who is diagnosed with mental health disorders, and with which mental health disorders people are diagnosed [49-51]. There are also inequities in who receives mental health care [52,53]. Much of the publicly available information and discourse about mental health contains inaccurate and stigmatizing information about mental health, and the existing research literature on mental health largely represents the perspectives of people who are White, are educated, are of high socioeconomic status, and speak English [54]. Far less information is available about the etiology of mental health concerns and effective assessments and interventions for populations that have been pushed to the margins. Training LLMs on existing data without appropriate safeguards and thoughtful human supervision and evaluation can, therefore, lead to problematic generation of biased content and disparate model performance for different groups [45,55-57] (of note, however, there is some evidence that clinicians perceive less bias in LLM-generated responses [58] relative to clinician-generated responses, suggesting that LLMs may have the potential to reduce bias compared to human clinicians).

LLMs should disseminate accurate, destigmatizing information about mental health and be trained to identify and combat stigma and discrimination. To do so, models need to be fine-tuned and evaluated for the mental health domain. Training models with data representative of the diverse populations being served is helpful, but new types of bias, such as semantic biases, may arise in LLMs [59]. Opportunities to train models to identify and exclude toxic and discriminatory language should be explored, both during the training of the underlying foundation models and during the domain-specific fine-tuning (see Keeling [59] for a discussion of the trade-offs of data filtration in this context) [45]. If LLMs perform differently for different groups or generate problematic or stigmatizing language during testing, additional model fine-tuning is required prior to deployment. Individuals developing LLMs should be transparent about the limitations of the training data, the approaches to data filtration and fine-tuning, and the populations for whom LLM performance has not been sufficiently demonstrated.

There is also hope that LLMs can be scaled to increase people’s access to mental health information, assessment, and treatment. LLMs have the potential to support delivery of mental health interventions in regions where access to mental health providers is limited and where significant barriers (eg, cost) exist. They can additionally help to personalize treatments to better fit people’s unique preferences, interests, identities, and language, hopefully improving treatment outcomes. LLMs may support increased access through more direct provision of mental health services, or LLMs can aid the expansion of the mental health workforce, training novice providers and community members in EBP at scale. There will undoubtedly be challenges in implementing and scaling LLMs globally. Revising and testing implementation frameworks for this new and evolving context and engagement in thoughtful public health and industry partnerships could all increase the likelihood that when mental health LLMs are scaled globally, implementation is sustained and best supports the populations most in need.

\subsection{Failing to Provide Mental Health Services Ethically}

A second risk is that LLMs will engage in unethical practices. When human mental health providers behave unethically, harm is done to patients and public trust is eroded [60]. LLMs will similarly do harm if they are not designed and implemented in consideration of and are not consistent with relevant ethical principles and standards when operating in the domain of mental health. Core ethical principles in the health care context include beneficence, nonmaleficence, justice, and autonomy [61]. Next, we highlight additional standards of ethical professional conduct that should apply when LLMs engage in mental health service provision (see the American Psychological Association Ethical Principles of Psychologists and Code of Conduct for parallel ethical principles and standards).

LLMs should operate within the boundaries of their competence and only engage in mental health tasks they have rigorously been proven to accomplish well. LLM developers should clearly communicate the limits and relevant evaluation results of LLMs, education should be provided to individuals about when it is and is not appropriate to use LLMs, and LLMs should withhold output when they are not competent in a task. LLM competence should be assessed and maintained over time. When competence is lacking in a certain domain, the LLM should no longer be deployed until the needed competence is gained (eg, via retraining and fine-tuning models with human validation).

Individuals should provide informed consent when interacting with mental health LLMs. They should be fully informed about the nature of mental health services they will receive and what role LLMs will have in that service. Information presented to individuals to help make decisions about consent should be understandable and include the possible risks and benefits of engaging with LLMs. Individuals should have the ability to choose not to consent to the use of LLMs in the direct provision of their mental health care, as well as the ability to withdraw their consent and opt out of the use of LLMs even if consent was initially given. As LLMs become further integrated into health care contexts, care should be taken to ensure that clients’ decisions to opt out of LLM involvement or to confine LLM involvement to less direct (eg, administrative) tasks do not limit their access to mental health care.

Confidentiality should be protected when individuals interact with LLMs to support their mental health. Individuals should be clearly informed about expectations for confidentiality. This should include information about the limits of confidentiality (eg, in the case of imminent risk for suicide), the foreseeable uses of information generated through engagement with LLMs, where and how their data are stored, and whether it is possible to delete their data. Policies related to data security should be strict and in line with relevant mental health data protection regulations [34]. Solutions such as developing on-device storage that does not require transmission of personal data [62] or systems with robust cloud-based encryption, pursuing LLMs that support compliance with relevant data protection laws (eg, Health Insurance Portability and Accountability Act [HIPAA]), and responsibly aggregating and deidentifying mental health data to fine-tune and test models all help to protect confidentiality.

Human mental health providers establish trusting relationships with those with whom they work and are obligated to ensure that the nature of the trusting provider-patient relationship does not lead to exploitation or harm. Appropriate trust is built through effective mental health assessment and treatment and, perhaps even more crucially, ethical practice. Trust should be evaluated through feedback from individuals engaged with LLMs. If and when trust is broken, this should be acknowledged and work should be done to repair trust. On the other hand, people may trust LLMs more than is warranted because of LLMs’ ability to produce humanlike natural language and to be trained to express emotion and empathy (this may especially be the case for individuals experiencing mental health concerns such as anxiety [63]) [64]. Unearned trust can have consequences, leading people to disclose personal information or trust content generated by LLMs even when it is not accurate. Education should be provided about the limits of LLMs and individuals should be cautioned against blanket trust in these models.

\subsection{Insufficient Reliability}

A third risk is that LLMs will not generate reliable or consistent output. When prompted to complete the same task or provide an answer to the same question multiple times, LLMs at times produce different responses [46,65]. Varied and creative output is a benefit of LLMs; however, the underlying response should be consistent even when articulated in different ways. Take for example an LLM repeatedly presented with a client’s description of depressive symptoms. The LLM should reliably reach the conclusion that the client meets the criteria for major depressive disorder even if this diagnostic conclusion is communicated to the client using different phrasing. Issues of low reliability of LLMs can erode trust and increase the possibility of harm, including leading some individuals to be misdiagnosed or to pursue treatments that are not best suited to their mental health concern.

LLM reliability should be measured and enhanced. Prompting approaches may help to improve LLM reliability. Self-consistency [66] and ensemble refinement [9] are strategies that sample multiple model answers to arrive at a more consistent response, improving model reliability [9]. Grounding models in data other than linguistic descriptions of symptoms (eg, objective behavioral or physiological signals) is another way of reducing variability in LLM performance, as words alone may not fully capture all of the necessary information to complete a given mental health task [67]. Finally, LLMs should not be deployed until they exceed prespecified thresholds of adequate reliability.

\subsection{Inaccuracy}

LLMs risk producing inaccurate information about mental health [46,68]. If LLMs are trained on data that contain inaccurate or outdated information, iatrogenic treatment options, or biased representations of mental health, that information can be reproduced by LLMs [45]. An additional consideration is that accuracy of LLM outputs has multiple dimensions and is not as simple to evaluate as answers to multiple-choice questions. Accuracy can be a function of how factual an answer is, how specific it is, or how devoid of irrelevant information it is. Generating inaccurate mental health information may be more damaging than no information, especially when it may be difficult for an individual to detect inaccuracies or inconsistencies (eg, about a complex mental health diagnosis).

Standards for accuracy should be defined a priori and should be high. When thresholds for LLM accuracy are not met, the risk of harm is too high and LLMs should not generate output. The accuracy of LLMs depends on the quality of data the model is trained and fine-tuned on [47,69,70]. LLMs should be adapted to the domain of mental health; models fine-tuned on mental health data perform better than models trained on non-domain-specific data [42] or general medical domains [16]. When data are limited, it is recommended that smaller but more variable data sets be prioritized over a larger single data set [19]). Training data should be highly curated, be grounded in authoritative and trusted sources, be specific to evidence-based health care, and represent diverse populations [46,58]. In mental health, the nature of consensus is continuing to evolve, and the amount of data available is continuing to increase, which should be taken into account when considering whether to further fine-tune models. Strategies such as implementing a Retrieval Augmented Generation system, in which LLMs are given access to an external database of up-to-date, quality-verified information to incorporate in the generation process, may help to improve accuracy and enable links to sources while also maintaining access to updated information. Accuracy of LLMs should be monitored over time to ensure that model accuracy improves and does not deteriorate with new information [45].

Measuring the accuracy of mental health LLMs is complex. It is not sufficient for models to merely outperform previous models. Rather, performance of LLMs should be compared with the performance of human clinicians, both of which should be compared against gold-standard, evidence-based care. When LLMs are tasked with mental health evaluation, their ability to predict scores on reliable and valid mental health assessments should be tested, and LLMs should meet human clinician performance in diagnostic accuracy. When LLMs are tasked with aiding mental health intervention delivery, their ability to detect, support, and engage in EBP is critical. Additional criteria to consider when evaluating the accuracy of LLMs include the level of agreement between human clinicians and LLMs, metrics of effect size rather than only statistical significance, and the balance of sensitivity and specificity in making diagnostic predictions.

LLMs should communicate confidence in the accuracy of generated output and limit or withhold output when confidence is lacking [58]. As an example, Med-PaLM 2’s accuracy improved when results were weighted based on confidence scores and when a cutoff threshold was set for confidence [20]. Communicating confidence in generated output and withholding output when confidence is low both help to enhance transparency and trust in LLMs’ ability to perform on mental health tasks and to limit potential harms associated with generating inaccurate information.

Prompt fine-tuning can boost LLM accuracy [9,19,58]. When applied to mental health, instruction fine-tuning improved performance of Mental-Alpaca relative to zero-shot and few-shot prompting and allowed Mental-Alpaca to reach a performance level across multiple mental health tasks (eg, identifying stress and classifying individuals as depressed or not based on Reddit posts) similar to that of Mental-RoBERTa, a task-specific model [19]. Prompting to concentrate on the emotional clues in text was also shown to improve ChatGPT performance on a variety of mental health-related tasks [71]. Conversely, however, instruction prompt fine-tuning can also increase inaccurate or inappropriate content [55]; thus, LLMs should continue to be evaluated for accuracy at all stages of prompt tuning.

\subsection{Lack of Transparency and Explainability}

LLMs risk generating output without being able to explain how they came to the decisions they did or without being able to identify the source of information used to generate the output [72]. There remains much that is not known about how LLMs generate reasoning for their responses and how sensitive these reasons are to context and prompting. It should be apparent when information is generated using LLMs, how LLMs were developed and tested, and whether LLMs are general-purpose or fine-tuned for the domain of mental health [46,58,68]. Additional steps to enhance transparency include explicitly telling individuals to exercise caution when interpreting or acting on LLM output and being clear about the bounds of LLMs’ competence [39].

Explainability, one aspect of transparency, was identified as a key priority by individuals engaged in mental health LLMs [39]. If asked to explain why they decided on a mental health diagnostic prediction or intervention, LLMs should explain what information was used to come to that decision. ChatGPT has been shown to be able to explain why an individual was classified as experiencing stress or depressive symptoms [71], and Med-PaLM 2 communicated why it predicted a particular symptom score and diagnosis [20]. Although LLMs are capable of producing plausible explanations through techniques such as chain-of-thought reasoning [73], more research is needed to ensure that explanations are internally consistent. Explainability is perhaps especially beneficial in the domain of mental health, as part of mental health assessment and intervention is communicating results of an evaluation or justification for an intervention to patients.

\subsection{Neglecting to Involve Humans}

There are risks associated with LLMs providing anonymous mental health services. Unlike mental health apps, where content can be highly curated, the content generated by LLMs is unpredictable. This makes interacting with LLMs more engaging, more appealing, and perhaps also more humanlike. However, it also increases the risk that LLMs may produce harmful or nontherapeutic content when tasked with independently providing mental health services. Legal and regulatory frameworks are needed to protect individuals’ safety and mental health when interacting with LLMs, as well as to clarify clinician liability when using LLMs to support their work or to clarify the liability of individuals and companies who develop these LLMs. There are ongoing discussions regarding the regulation of LLMs in medicine [74-76] that can inform how LLMs can support mental health while limiting the potential for harm and liability.

Humans should be actively involved in all stages of mental health LLM development, testing, and deployment. For mental health LLMs to be effective, rigorous, and ongoing, human supervision and input are needed (see Figure 4) [46]. Reinforcement learning through human feedback can improve model accuracy and uncover problematic LLM responses [14,42]. This feedback should be obtained from individuals who reflect the diverse populations the LLM aims to help, including members of the public, patients, and human clinicians [9,14,34,58,68,77,78]. Their input should be leveraged to identify and correct biases, to ensure generated content is inclusive, culturally appropriate, and accurate, and to reduce the likelihood of harm. Particularly important is prioritizing the perspectives of individuals at heightened risk for mental health concerns (eg, sexual and gender minorities) and individuals with lived experience with mental health concerns. These individuals should play a central role in co-defining the role LLMs will play in mental health care and in co-designing tools that leverage LLMs. Practically, use cases should focus on opportunities to support and augment provider care. As just one example, LLMs may have a role in suggesting language used in clinical notes, but clinicians should have the final say in whether they adopt those suggestions or not.

\begin{figure}[H]
  \centering
  \includegraphics[width=0.8\textwidth]{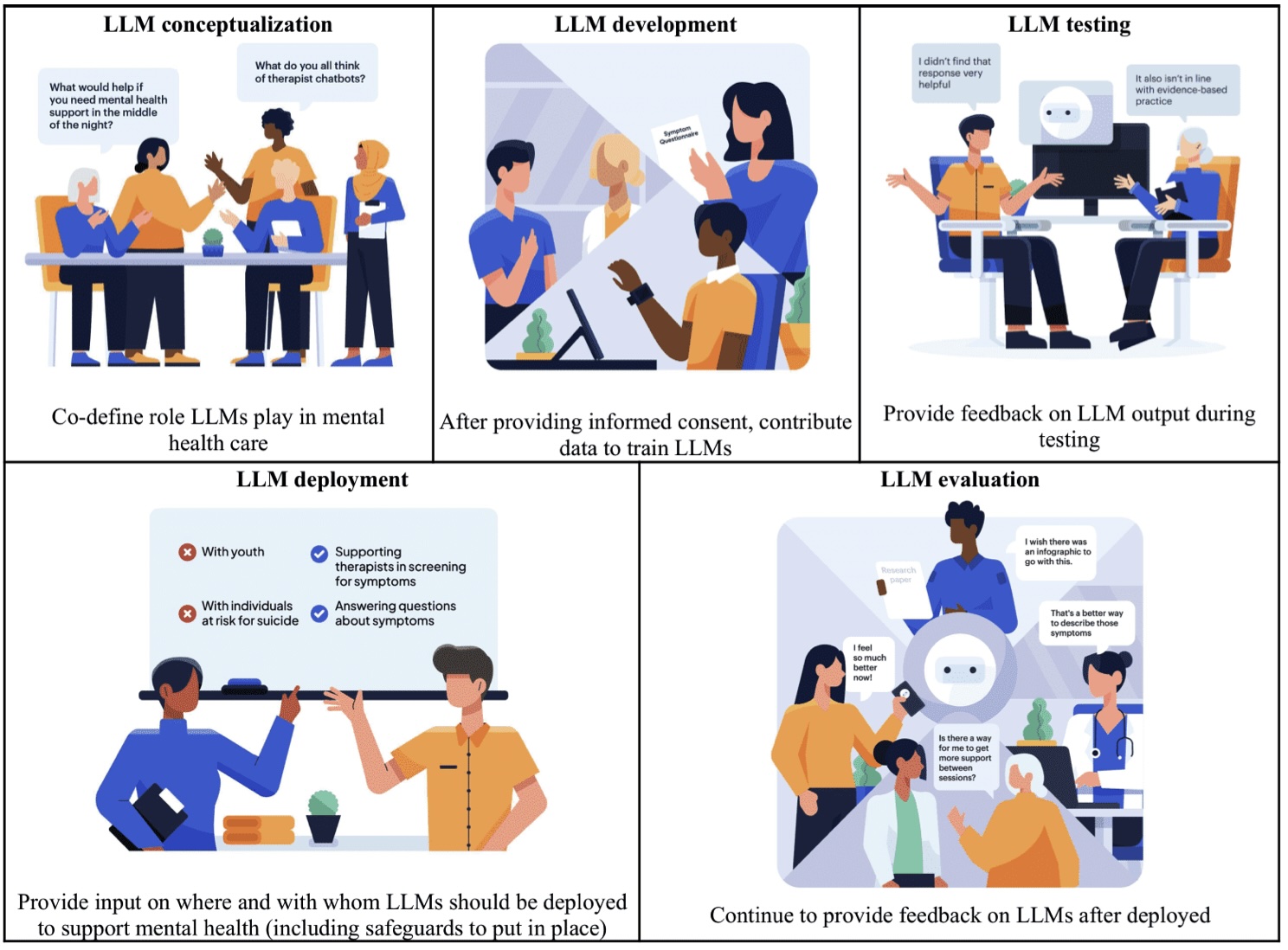}
  \caption{Examples of human involvement across all stages of LLM development through deployment and evaluation. LLM: large language model.}
  \label{fig:myimage4}
\end{figure}

\section{Conclusions}

The need for mental health services is pressing, and the potential of LLMs to expand access to information about mental health and to mental health care is great. LLMs are advancing rapidly and have been applied across mental health education, assessment, and intervention. Especially promising is the potential for LLMs to provide mental health education and assessment—tasks that are well aligned with LLM strengths. LLMs have made exceptional progress in related tasks such as answering medical questions and assessing medical conditions, reaching and in some cases exceeding the performance of human clinicians. Greater caution is warranted when applying LLMs to mental health intervention, but there is also cause for optimism that LLMs could eventually help to support or augment human provision of mental health treatments. Additional research is needed in testing LLMs’ ability to deliver or train providers in empirically supported treatments, to responsibly adapt approaches for youth and marginalized populations, to build appropriate rapport, and to detect risk for high-acuity mental health concerns for progress to be made in these areas.

Critical to effectively engaging in mental health care tasks is fine-tuning LLMs specifically for the domain of mental health and the prioritization of equity, safety, EBP, and confidentiality. No widely used, general-purpose LLM has been fine-tuned for mental health, trained on evidence-based mental health content, or sufficiently tested on mental health-related tasks. When LLMs are developed specifically for mental health, tested to ensure adherence with EBP, and aligned with the goals of people with lived experience with mental health concerns and those who have expertise in mental health care, there is great hope that they will expand access to evidence-based mental health information and services. Investing in developing, testing, and deploying mental health LLMs responsibly has the potential to finally reverse rising global mental health rates and to improve the mental health of the millions of people in need of mental health support.

\section{Acknowledgments}

We acknowledge and thank Michael Howell, MD, MPH; Bakul Patel, MSEE, MBA; Matthew Thompson, DPhil, MPH; Joseph Dooley, MPA; and David Steiner, MD, PhD, for reviewing and providing helpful feedback on this paper.

\section{Conflicts of Interest}

RAS, MJM, DJM, and MJB are employees of Google and receive monetary compensation from Google and equity in Google's parent company, Alphabet. HRL and SBR are employees of, and receive compensation from, Magnit and are contracted for work at Google. In addition, MJB is a shareholder in Meeno Technologies, Inc and The Orange Dot (Headspace Health). RAS is a shareholder in Lyra Health and Trek Health, and she consults with Understood.

\end{document}